\documentclass[11pt]{article}

\usepackage{acl}
\usepackage{times}
\usepackage{latexsym}

\usepackage[T1]{fontenc}
\usepackage[utf8]{inputenc}

\usepackage{microtype}
\usepackage{amsmath,amssymb,amsthm}

\usepackage{inconsolata}

\usepackage{graphicx}
\usepackage{booktabs}

\usepackage{xcolor}
\usepackage{tcolorbox}
\usepackage{enumitem}
\usepackage{listings}
\usepackage{authblk} % 处理多单位和作者排版的核心宏包
\title{Hindsight Credit Assignment over Trees for Agentic RL}
\author{
  \textbf{Penghong Zhao}$^{1*}$, 
  \textbf{Yihao Chen}$^{1*}$, 
  \textbf{Maoheng Yuan}$^{2*}$, 
  \textbf{Yunjian Zhang}$^1$, 
  \textbf{Abudukelimu Wuerkaixi}$^1$, \\
  \textbf{Haoran Xu}$^1$, 
  \textbf{Man Lv}$^1$, 
  \textbf{Yaolong Wang}$^1$, 
  \textbf{Zhijie Sun}$^1$,
  \textbf{Guohua Liu}$^1$, 
  \textbf{Yuewei Zhang}$^{1\dagger}$ \\
  $^1$Alibaba Cloud Computing \quad $^2$Fudan University \\
 
  \texttt{\{zhaopenghong.zph, cyh535471, liyou.zyw\}@alibaba-inc.com} \\
  
  \texttt{mhyuan24@m.fudan.edu.cn}
}

\begin{document}
\maketitle

\begingroup
\renewcommand\thefootnote{}\footnote{$^*$Equal contribution.}%
\addtocounter{footnote}{0}%
\footnote{$^\dagger$Corresponding author.}%
\endgroup

\begin{abstract}
Reinforcement learning for long-horizon LLM agents faces a fundamental dilemma: successful experiences are essential for policy improvement, yet limited policy capabilities make such experiences difficult to discover. Sparse outcome rewards compound this challenge, leaving sampled trajectories with little relative
feedback for learning. In this paper, we propose TriGRPO, a tree-based reinforcement learning framework that turns scarce successes into exploration and learning signals through hindsight credit assignment within and across rollout trees. TriGRPO constructs rollout trees through hindsight-guided branching: when an ongoing trajectory reaches a matched decision state, a successful action discovered during rollout initiates a sibling branch that continues autonomously, enabling further exploration and parent--sibling comparisons. Three complementary groups organize policy optimization over these trees: root groups capture within-prompt relative returns, node groups assign credit from parent--sibling outcome differences, and cross groups supplement sparse feedback with historical performance on related tasks. Experimental results demonstrate consistent gains in task performance and rollout efficiency across model scales, highlighting structured hindsight credit assignment as an effective approach to long-horizon agent learning.
\end{abstract}

\section{Introduction}
\label{sec:introduction}

Large language models (LLMs) increasingly act as agents that sustain progress through sequential decisions and environmental feedback. Reinforcement learning (RL) improves these capabilities from interaction outcomes \cite{wang2026beyond,zhang2026resource,DBLP:conf/iclr/YangLWCLZCJ26}, but must turn behaviors discovered during exploration into reliable strategies for independent task solving.

\begin{figure}[t]
  \centering
  \includegraphics{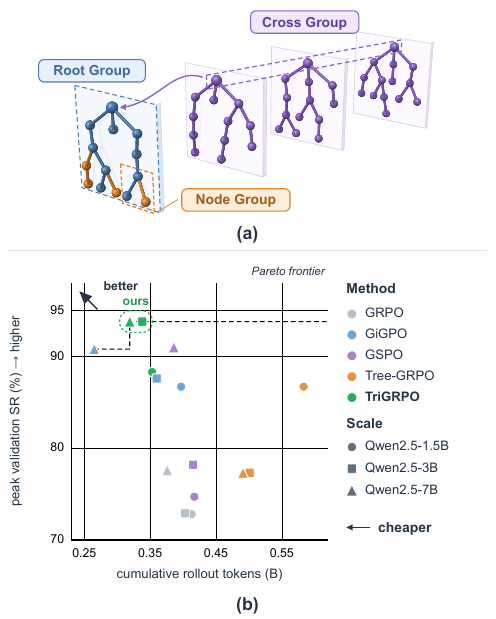}
\caption{TriGRPO enhances agentic RL through hindsight credit assignment. (a) Credit assignment across trees, within a tree, and at branch nodes. (b) Superior Pareto frontier of success versus cost.}
  \label{fig:firstfig}
\end{figure}

Existing approaches compare sampled trajectories: GRPO \citep{shao2024deepseekmath} uses within-group returns without a learned critic, while GSPO \citep{zheng2025gspo} refines sequence-level optimization. For multi-turn agents, GiGPO derives step-level credit from recurring states \citep{feng2025gigpo}, and Tree-GRPO compares prefix-sharing continuations \citep{ji2025treegrpo}. CAST uses game-solver values for turn-level supervision \citep{wang2026cast}; PVPO addresses degenerate groups through difficulty filtering and pre-estimated baselines \citep{feng2026pvpo}.

Yet autonomous exploration limits the experience available for learning. Under sparse rewards, failures can dominate sampling without providing within-group relative signals. The policy needs successful experience to improve, yet its limited capability makes that experience difficult to obtain. Occasional successes offer a foothold: repeating useful decisions enables deeper exploration \citep{oh2018sil}, while replay supports further updates \citep{zhan2026exgrpo}. We seek to make scarce successes guide new interactions, helping other trajectories build on discovered behaviors and learn to complete tasks independently.

% We propose \textbf{TriGRPO} (\underline{Tri}ple \underline{G}roup \underline{R}elative \underline{P}olicy \underline{O}ptimization), coupling hindsight-guided exploration with hierarchical credit assignment. Successful decisions guide further exploration, while three comparison scales evaluate the resulting experience to support autonomous task solving. Under exact matching of sufficient decision states, our analysis shows that success-conditioned action replacement yields a nonnegative expected gain in task success probability, motivating state-aligned branching at action disagreements.
We propose \textbf{TriGRPO} (\underline{Tri}ple \underline{G}roup \underline{R}elative \underline{P}olicy \underline{O}ptimization), a unified framework that operationalizes hindsight successes into multi-scale relative optimization. Rather than passive offline replay, TriGRPO exploits discovered successes strictly online along two dimensions: within-tree hindsight branching to actively guide exploration, and across-tree hindsight credit assignment to enrich sparse learning signals. 
Under exact matching of sufficient decision states, our analysis shows that success-conditioned action replacement yields a nonnegative expected gain in task success probability, motivating state-aligned branching at action disagreements (Figure~\ref{fig:firstfig}).

% TriGRPO takes candidate actions from successful unguided trajectories for the same prompt in the current batch. When a deviating trajectory reaches a matching observable state, a sibling executes a candidate action and continues autonomously alongside the unchanged parent. Unguided trajectories receive \emph{root credit} from within-prompt comparisons and \emph{cross credit} against related tasks' historical performance, supplementing relative feedback when current returns offer little differentiation. Guided continuations receive only \emph{node credit} from parent--sibling outcome differences. Injected actions receive only behavior cloning, and autonomous decisions receive reinforcement learning with their corresponding credit. No external demonstrations or preliminary sampling are required. 
TriGRPO structures multi-scale credit assignment around three comparison groups derived from this hindsight mechanism. At the prompt level, standard unguided rollouts form a root group evaluated against within-prompt returns. Within each tree, when an ongoing rollout reaches an observable state previously visited by a completed successful trajectory, a new branch splits off by executing the corresponding hindsight action and continues autonomously, forming a node group; this branched trajectory receives only node credit from parent--sibling outcome differences.
Across trees, unguided rollouts reference related tasks' historical performance to form a cross group without generating additional rollouts, receiving cross credit against this historical benchmark. Injected actions are supervised solely with behavior cloning, while autonomous decisions are optimized via reinforcement learning with their corresponding credits. No external demonstrations or preliminary sampling are required.
% On ALFWorld, TriGRPO reaches 88.3\%, 93.8\%, and 93.8\% success with Qwen2.5-1.5B, 3B, and 7B, exceeding the strongest baselines by 1.6, 6.2, and 2.8 points and using 32.5--39.4\% fewer rollout tokens than Tree-GRPO. Retrieval-augmented QA, ablations, and difficulty-stratified analyses further assess performance and component contributions.
Our contributions are summarized as follows:
\begin{itemize}
    \item We establish a theoretical guarantee that success-conditioned action replacement yields nonnegative expected gains, which motivates an online hindsight-branching strategy requiring no offline replay, external demonstrations, or auxiliary sampling overhead.
    % \item We propose TriGRPO, a unified framework that projects hindsight successes both within and beyond prompts, driving credit assignment over branches and trees to structure policy optimization across three groups.
    \item We propose \textbf{TriGRPO}, a unified framework that translates hindsight experience into complementary policy optimization signals across three comparison groups: root groups for within-prompt relative credit, node groups for branch-level credit, and cross groups for historical feedback across related tasks.
    \item Extensive experiments demonstrate that TriGRPO improves success rates by up to 6.2\% over state-of-the-art baselines while reducing rollout token consumption by up to 39.4\%.
\end{itemize}

\section{Related Work}
\label{sec:related}

\noindent\textbf{Group-Based Policy Optimization and Credit Assignment.}
GRPO \citep{shao2024deepseekmath} replaces the learned critic in PPO \citep{schulman2017ppo} with group-relative baselines, while variants such as DAPO \citep{yu2025dapo}, Dr.~GRPO \citep{liu2025drgrpo}, and GSPO \citep{zheng2025gspo} improve optimization stability. Other studies leverage reward variance or auxiliary rollouts to reshape feedback \citep{jiang2025vcrl,jiang2026dvao,feng2026pvpo}. For step-level credit, GiGPO \citep{feng2025gigpo} clusters naturally recurring anchor states, and CAST \citep{wang2026cast} relies on external solver values. In contrast, TriGRPO requires no external solvers; it couples a cross-batch historical baseline with localized parent--sibling counterfactual comparisons at matched decision states.

\noindent\textbf{Tree-Based Rollouts and Exploration.}
Tree topologies support both inference planning \citep{yao2023tree,zhou2023language} and policy optimization. TreeRL \citep{hou2025treerl} and Tree-GRPO \citep{ji2025treegrpo} employ tree rollouts with prefix sharing and hierarchical advantage estimation. Unlike Tree-GRPO's unguided random expansion, TriGRPO initiates branches deterministically when an active rollout matches a state banked by a successful trajectory from the same prompt, concentrating compute where verified alternative actions are known. We evaluate against GiGPO and Tree-GRPO as our primary baselines for step-level credit and tree exploration, respectively.

\noindent\textbf{Hindsight and Experience Reuse.}
Hindsight learning extracts training signals from past rollouts \citep{andrychowicz2017her}. In LLMs, ECHO \citep{hu2025echo} and HSL \citep{li2026hsl} perform post-hoc goal relabeling to construct synthetic demonstrations, while SIL \citep{oh2018sil} and ExGRPO \citep{zhan2026exgrpo} replay prioritized successes. Rather than altering goals or replaying offline trajectories, TriGRPO applies hindsight online at the action level, injecting verified actions into concurrent rollouts to compute counterfactual policy-gradient advantages in real time.

% Originating from Hindsight Experience Replay \citep{andrychowicz2017her}, hindsight methods extract learning signals from unrewarded trajectories. Recent works adapt this paradigm to LLM agents: ECHO \citep{hu2025echo} prompts an LLM to rewrite failed trajectories toward alternative goals reachable under observed transitions to guide subsequent attempts, while HSL \citep{li2026hsl} employs an auxiliary LLM to relabel a completed trajectory with the goals it actually achieved and fine-tunes on the relabeled data. A defining characteristic of these methods is their \emph{post-hoc, goal-relabeling} design: hindsight is enacted offline after episode termination to synthesize supplementary supervision. In contrast, TriGRPO integrates hindsight \emph{online at the action level}: rolling trajectories dynamically query an active bank of successful transitions, instantiating paired counterfactual branches at matched states during the rollout itself. This operationalizes hindsight as an engine for real-time policy-gradient credit assignment rather than offline data augmentation, rendering our framework conceptually orthogonal and distinct from post-hoc relabeling pipelines.

\begin{figure*}[ht]
  \centering
  \includegraphics[width=\textwidth]{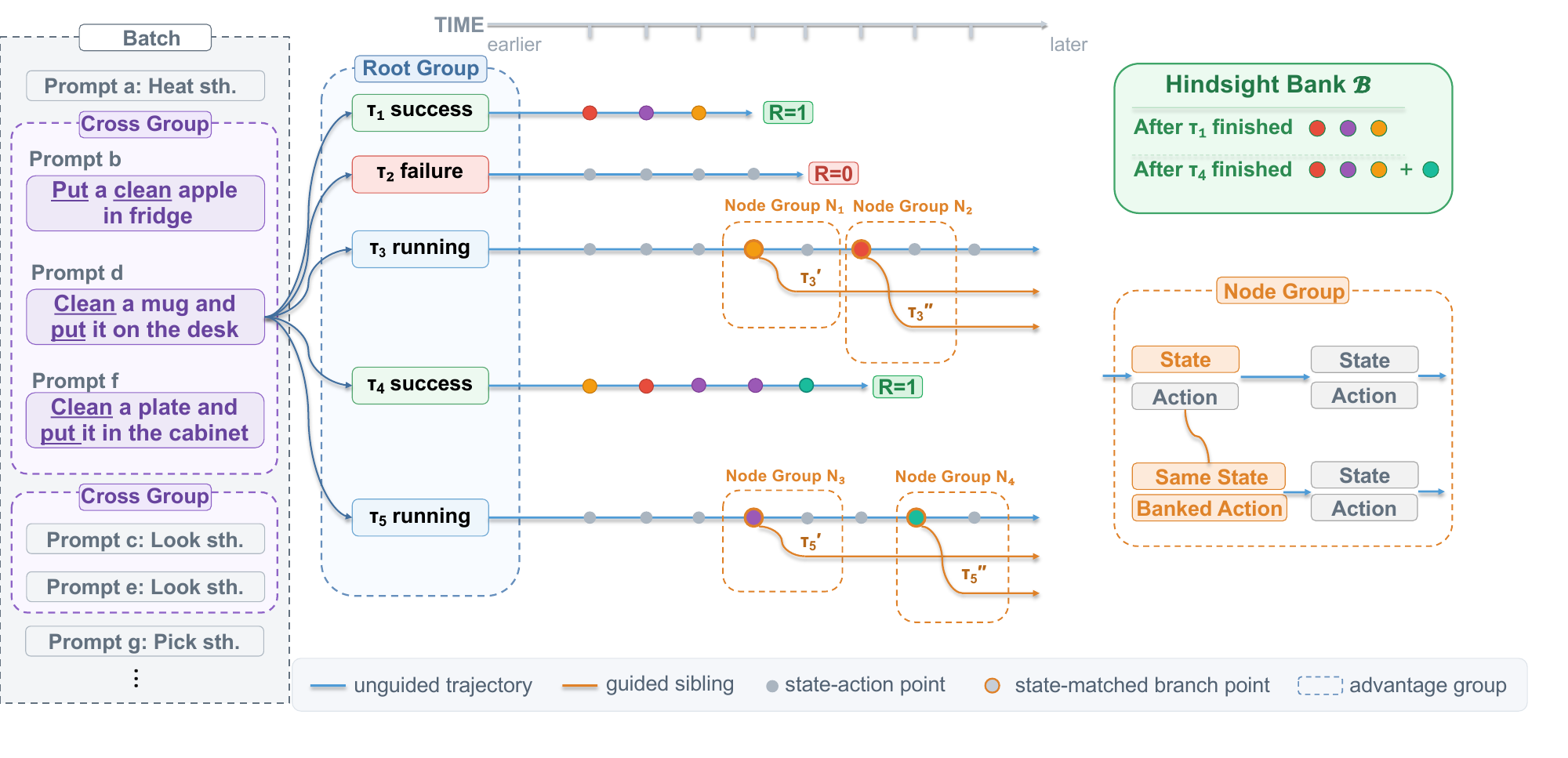}
 \caption{Overview of the TriGRPO training pipeline. Tree-structured rollouts generate $R$ root trajectories per prompt, expanded into prefix-sharing trees. 
 During training, successful trajectories populate the hindsight bank in real time. 
 Upon detecting a matched state, TriGRPO creates a guided branch by injecting the hindsight action, forming a node group.
 Prompts sharing the same task type form a cross group to enable cross-prompt credit assignment, propagating hindsight experience across tasks. 
 Once batch rollouts are completed, a triple-group advantage estimator assigns node, root, and cross credits for policy updates.}
  \label{fig:pipeline}
\end{figure*}

\section{Method}
\subsection{Preliminaries}
\label{sec:preliminaries}

We consider a multi-turn agent interacting with an environment to accomplish a task specified by a prompt $x$. At each step $t$, the agent receives an observation $o_t$ and selects an action $a_t$ according to a policy $\pi_\theta(\cdot\mid h_t)$, where $h_t=(x,o_1,a_1,\ldots,o_t)$ denotes the interaction history. This process generates a trajectory $\tau=(o_1,a_1,\ldots,o_T,a_T)$ until termination, with a binary outcome reward $R(\tau)\in\{0,1\}$ indicating task success or failure. For each prompt $x$ we sample a group of $N$ trajectories, $\mathcal{G}_x=\{\tau_i\}_{i=1}^{N}$.

% We adopt a group-based policy optimization framework, where a group of $N$ trajectories $\mathcal{G}=\{\tau_i\}_{i=1}^{N}$ is sampled for each prompt $x$. The group-relative advantage of trajectory $\tau_i$ is computed as
% \begin{equation}
%     A_i =
%     \frac{R(\tau_i)-\bar{R}_{\mathcal{G}}}
%     {\sigma_{\mathcal{G}}},
%     \label{eq:grpo-advantage}
% \end{equation}
% where $\bar{R}_{\mathcal{G}}$ and $\sigma_{\mathcal{G}}$ denote the mean and standard deviation of rewards within the group, respectively. This formulation enables policy optimization through relative outcome comparisons without requiring a learned critic.

\subsection{Theoretical Analyses}
\label{sec:theoretical_analysis}

Let $G$ denote the event of eventual task success and $s$ a sufficient decision state that captures all information relevant to future interactions, including the environment state, policy-visible history, and remaining horizon. Under a fixed policy $\pi$, we define the state- and action-level success probabilities as
\begin{equation}
\small
\begin{aligned}
    h(s) &:= \Pr_\pi(G\mid S=s),\\
    q(s,a) &:= \Pr_\pi(G\mid S=s,A=a),
\end{aligned}
\label{eq:success_values}
\end{equation}
where the trajectory continues under $\pi$ after action $a$; by the law of total probability, $h(s)=\mathbb{E}_{A\sim\pi(\cdot\mid s)}[q(s,A)]$.

For any state satisfying $h(s)>0$, Bayes' rule gives the action distribution conditioned on eventual success:
\begin{equation}
\small
\pi^+(a\mid s)
:=\Pr_\pi(A=a\mid S=s,G)
=\frac{\pi(a\mid s)q(s,a)}{h(s)}.
\label{eq:success_conditioned_action}
\end{equation}
We formalize this property in the following proposition.

\noindent\textbf{Proposition 1 (Success-Conditioned Action Advantage).}
Consider a decision state $s$ with $h(s)>0$. Let $A^+\sim\pi^+(\cdot\mid s)$ denote an action sampled from the success-conditioned distribution, and $A^0\sim\pi(\cdot\mid s)$ an action sampled from the original policy. When both actions are evaluated with subsequent continuation under $\pi$, the expected success-probability gain of $A^+$ over $A^0$ satisfies
\begin{equation}
\small
    \mathbb{E}[q(s,A^+)-q(s,A^0)] =
    \frac{
    \operatorname{Var}_{A\sim\pi(\cdot\mid s)}
    [q(s,A)]
    }{h(s)}
    \geq 0.
\label{eq:replacement_gain}
\end{equation}

\noindent\emph{Proof.}
Substituting Eq.~\eqref{eq:success_conditioned_action}, we obtain
\begin{equation}
\small
    \mathbb{E}[q(s,A^+)]=\frac{
    \mathbb{E}_{A\sim\pi(\cdot\mid s)}
    [q(s,A)^2]
    }{h(s)},
    \mathbb{E}[q(s,A^0)]=h(s).
\end{equation}
Therefore,
\begin{equation}
\small
\begin{aligned}
    &\mathbb{E}[q(s,A^+)-q(s,A^0)]\\
    &\quad =
    \frac{
    \mathbb{E}[q(s,A)^2]
    -\mathbb{E}[q(s,A)]^2
    }{h(s)} =
    \frac{
    \operatorname{Var}[q(s,A)]
    }{h(s)}
    \geq 0,
\end{aligned}
\end{equation}
where the expectations and variance in the last two lines are taken over $A\sim\pi(\cdot\mid s)$.
\hfill$\square$

\paragraph{Implications for Branching.}
Proposition 1 motivates three principles for success-guided branching:

\begin{enumerate}
    \item \textbf{Where to branch: State alignment.}
    The gain in Eq.~\eqref{eq:replacement_gain} requires a shared decision state, motivating state alignment to identify valid branching points.

    \item \textbf{When to branch: Action disagreement.}
    Let $D=\{A^+\neq A^0\}$. Since identical actions contribute no gain,
    \begin{equation}
    \small
        \mathbb{E}\!\left[
        \bigl(q(s,A^+)-q(s,A^0)\bigr)
        \mathbf{1}_{D}
        \right]
        =
        \frac{
        \operatorname{Var}_{A\sim\pi(\cdot\mid s)}
        [q(s,A)]
        }{h(s)}.
    \label{eq:disagreement_gate}
    \end{equation}
    Thus, branching can focus on action disagreements without losing the unconditional expected replacement gain.

    \item \textbf{How to branch: Prefix-preserving action injection.}
    At a matched decision state, we preserve the recipient's prefix and inject the successful action to create an alternative continuation, so that action replacement occurs at the same pre-action state (consistent with Eq.~\eqref{eq:replacement_gain}) while avoiding redundant prefix generation. Let $C_p$ denote the cost of generating a common prefix and $C_c$ that of a subsequent continuation. Generating $K$ sibling trajectories independently incurs $K(C_p+C_c)$, whereas branching from the shared prefix requires $C_p+KC_c$, yielding an idealized local compute saving of
    \begin{equation}
        \Delta C = (K-1)C_p.
        \label{eq:prefix_cost}
    \end{equation}
    In practice, this upper-bound economy amortizes prefix generation over active hindsight continuations while avoiding redundant rollouts of shared decision histories.
\end{enumerate}

The theoretical gain holds under exact state matching and success-conditioned action sampling, while the computational benefit arises from amortizing shared-prefix generation costs.

\subsection{TriGRPO}
\label{sec:trigrpo}

Building on Section~\ref{sec:theoretical_analysis}, we propose TriGRPO, integrating three components: (i) prefix-sharing tree rollouts, (ii) hindsight branching guided by successful actions, and (iii) hindsight credit assignment through root-, cross-, and node-level advantages.

\paragraph{Tree-structured rollouts.}
Rather than sampling $N$ trajectories independently for each prompt $x$, we generate $R$ root trajectories and expand sampled non-leaf nodes into prefix-sharing trees~\citep{ji2025treegrpo}. We denote by $\mathcal{G}_x$ the set of all generated trajectories (root and branch) for prompt $x$. Prefix sharing eliminates redundant generation and enables comparisons at multiple levels of the trajectory hierarchy.

\paragraph{Hindsight Branching.}
Following Eqs.~\eqref{eq:replacement_gain} and~\eqref{eq:disagreement_gate}, TriGRPO maintains a dynamic success bank $\mathcal{B}$ of successful trajectories with $R(\tau)=1$ from the current prompt. When an active trajectory $\tau_i$ reaches a decision state $s$ that matches a state $s'$ visited by a trajectory in $\mathcal{B}$, and the two actions disagree ($a^+ \neq a_i$, where $a_i$ is the action $\tau_i$ would have taken at $s$), we spawn a counterfactual sibling branch $\tau_j$ that inherits $\tau_i$'s prefix up to $s$, substitutes the banked action $a^+ \in \mathcal{B}$, and continues autonomously under $\pi_\theta$. We denote unguided trajectories by $\mathcal{U}$ and guided branches by $\mathcal{V}$; each pair $(\tau_i, \tau_j)$ forms a \emph{node group} isolating the causal effect of the action at $s$ (Figure~\ref{fig:pipeline}).

\paragraph{Three-level advantage estimator.}
We capture learning signals at three granularities.

\emph{Root-level advantage.} For each unguided trajectory $\tau_i \in \mathcal{U}$:
\begin{equation}
  A_{\text{root}}(\tau_i) = \frac{R(\tau_i) - \bar{R}_x}{\sigma_x},
  \label{eq:root-adv}
\end{equation}
where $\bar{R}_x$ and $\sigma_x$ are the mean and standard deviation of returns over all trajectories in $\mathcal{G}_x$.

\emph{Cross-sample advantage.} To prevent vanishing advantages when all trajectories in a group receive identical rewards, we maintain historical baselines at both task and task-type levels. Let $\mu_f(t)$ and $\mu_c(t)$ denote EMA rewards for task $t$ and its task type $c$; after each step they move toward the corresponding batch mean returns $\bar{R}_t$ and $\bar{R}_c$ with smoothing coefficient $\alpha$:
\begin{equation}
 \mu_k(t) \leftarrow (1-\alpha)\,\mu_k(t) + \alpha\,\bar{R}_k, \quad \forall k \in \{f, c\}.
  \label{eq:ema-update}
\end{equation}
To balance task-specific precision against task-type stability, we apply Bayesian shrinkage toward the coarse prior:
\begin{equation}
  \mu(t) = \frac{n_t \cdot \mu_f(t) + \kappa \cdot \mu_c(t)}{n_t + \kappa},
  \label{eq:ema-baseline}
\end{equation}
where $n_t$ is the number of previously observed root trajectories for task $t$ and $\kappa$ controls the prior strength. The cross-sample advantage is then
\begin{equation}
  A_{\text{cross}}(\tau_i) = \frac{R(\tau_i) - \mu(t)}{\sigma_r},
  \label{eq:cross-adv}
\end{equation}
where $\sigma_r>0$ is a reward normalization factor; it is applied strictly to unguided trajectories $\tau_i \in \mathcal{U}$, excluding guided branches from historical estimation.

\emph{Node-level advantage.} For each node group $(\tau_i, \tau_j)$:
\begin{equation}
  A_{\text{node}}(\tau_j) = R(\tau_j) - R(\tau_i).
  \label{eq:node-adv}
\end{equation}
Following C3~\citep{chen2026c3}, instantiated here as an unguided--guided trajectory pair $(\tau_u, \tau_v) \in \mathcal{U} \times \mathcal{V}$, this isolates the counterfactual gain of action $a^+$ at decision state $s$. It needs no variance normalization: each node group holds exactly two binary-reward trajectories, so the difference is naturally bounded in $\{-1,0,1\}$ and any residual scale is absorbed by $\lambda_{\text{node}}$.

\paragraph{Combined advantage and optimization.}
The unified advantage is
\begin{equation}
  \hat{A}(\tau) =
  \begin{cases}
    A_{\text{root}}(\tau) + \lambda_{\text{cross}}\, A_{\text{cross}}(\tau) & \text{if }\tau \in \mathcal{U},\\[2pt]
    \lambda_{\text{node}}\, A_{\text{node}}(\tau) & \text{if }\tau \in \mathcal{V},
  \end{cases}
  \label{eq:combined-adv}
\end{equation}
where $\lambda_{\text{cross}}$ and $\lambda_{\text{node}}$ control the relative contribution of each credit level. 
To further stabilize the policy around successfully injected actions and accelerate early training, we introduce an auxiliary behavior cloning (BC) loss.
The policy is then optimized via
\begin{equation}
  \mathcal{L} = \mathcal{L}_{\text{RL}} + \eta(t) \cdot \mathcal{L}_{\text{BC}},
  \label{eq:objective}
\end{equation}
where $\mathcal{L}_{\text{RL}}$ uses $\hat{A}(\tau)$ within KL-regularized GRPO and $\mathcal{L}_{\text{BC}}$ supervises the injected actions $a^+$ on guided branches $\mathcal{V}$. The BC weight $\eta(t)$ is held constant early and then decayed to a small floor, gradually shifting emphasis from imitation to reinforcement learning.

\section{Experimental Results}

\subsection{Setup}
\label{sec:experiment_setup}

Experiments are conducted on two interactive benchmarks: \textbf{ALFWorld}~\citep{shridhar2021alfworld} ($n{=}128$ test tasks) and \textbf{SearchQA}~\citep{jin2025search} (51{,}713 test questions spanning 7 benchmarks). We report the peak validation success rate (SR) averaged over 3 random seeds, alongside cumulative token consumption up to the peak step to quantify sampling efficiency. Detailed environment configurations are provided in Appendix~\ref{app:tools} and~\ref{app:eval_protocol}.

We compare against: (1)~\textbf{GRPO} (root-only), (2)~\textbf{Tree-GRPO} \citep{ji2025treegrpo}, (3)~\textbf{GiGPO} \citep{feng2025gigpo}, and (4)~\textbf{GSPO} \citep{zheng2025gspo}. All baselines share the training configuration and evaluation protocol above; reproduction details are in Appendix~\ref{app:baselines}.

Unless otherwise noted, all runs use a learning rate of $1\times10^{-6}$ with a constant schedule. On ALFWorld the tree rollout uses $R{=}8$ root trajectories per prompt with up to $L{=}12$ leaves and a training batch of 16 prompts per step for 150 steps; on SearchQA the budget is reduced to $R{=}3$ roots with up to $L{=}5$ leaves and a batch of 256 prompts per step. Complete hyperparameters are in Appendix~\ref{app:hyperparams}.

\begin{table*}[ht]
\centering
\small
\begin{tabular}{l cccccc cc}
\toprule
\textbf{Method} & pick & look & clean & heat & cool & pick2 & \textbf{All} & \textbf{Tokens} \\
\midrule
\multicolumn{9}{l}{\textit{Qwen2.5-1.5B-Instruct}} \\
\midrule
Base (no RL) & 5.9 & 5.5 & 3.3 & 9.7 & 4.2 & 0.0 & 4.1 & -- \\
GRPO & 85.3 & 53.7 & 84.5 & 78.2 & 59.7 & 53.5 & 72.8 & 0.413B \\
GiGPO & 94.4 & 67.5 & 94.8 & 94.4 & 79.8 & 76.4 & 86.7 & 0.397B \\
GSPO & 89.3 & 55.3 & 87.7 & 83.2 & 78.5 & 54.0 & 74.7 & 0.417B \\
Tree-GRPO & \textbf{97.1} & \textbf{73.3} & 95.5 & 94.4 & \textbf{95.2} & 50.0 & 86.7 & 0.583B \\
TriGRPO (ours) & 92.3 & 66.7 & \textbf{100.0} & \textbf{94.7} & 76.5 & \textbf{85.7} & \textbf{88.3} & \textbf{0.353B} \\
\midrule
\multicolumn{9}{l}{\textit{Qwen2.5-3B-Instruct}} \\
\midrule
Base (no RL) & 36.3 & 20.7 & 6.9 & 12.8 & 9.1 & 5.5 & 15.2 & -- \\
GRPO & 91.4 & 63.6 & 70.6 & 71.4 & 64.0 & 76.2 & 72.9 & 0.403B \\
GiGPO & 90.1 & 78.9 & 85.6 & 92.1 & \textbf{95.2} & 83.8 & 87.6 & 0.360B \\
GSPO & 92.6 & 71.4 & 88.5 & 60.0 & 78.1 & 78.3 & 78.2 & 0.415B \\
Tree-GRPO & \textbf{96.7} & 73.3 & 88.9 & 46.2 & 75.0 & 57.9 & 77.3 & 0.501B \\
TriGRPO (ours) & 96.6 & \textbf{86.7} & \textbf{95.8} & \textbf{100.0} & 91.3 & \textbf{88.2} & \textbf{93.8} & \textbf{0.338B} \\
\midrule
\multicolumn{9}{l}{\textit{Qwen2.5-7B-Instruct}} \\
\midrule
Base (no RL) & 33.4 & 21.6 & 19.3 & 6.9 & 2.8 & 3.2 & 14.8 & -- \\
GRPO & 90.8 & 66.1 & 89.3 & 74.7 & 72.5 & 64.7 & 77.6 & 0.376B \\
GiGPO & 97.7 & 82.7 & 98.8 & 83.7 & 89.3 & 79.2 & 90.8 & \textbf{0.265B} \\
GSPO & \textbf{100.0} & \textbf{100.0} & \textbf{100.0} & 92.9 & \textbf{96.8} & 56.0 & 91.0 & 0.386B \\
Tree-GRPO & 90.3 & 92.9 & 95.7 & 50.0 & 77.8 & 53.8 & 77.3 & 0.491B \\
TriGRPO (ours) & \textbf{100.0} & 94.4 & \textbf{100.0} & \textbf{93.8} & 76.5 & \textbf{91.7} & \textbf{93.8} & 0.319B \\
% midrule
% \multicolumn{9}{l}{\textit{Llama-3.2-3B-Instruct}} \\
% \midrule
% Base (no RL) & 42.4 & 11.1 & 10.3 & 0.0 & 9.1 & 0.0 & 12.2 & -- \\
% GRPO & 83.3 & 70.0 & 64.3 & 46.2 & 37.5 & 50.0 & 58.6 & 0.577B \\
% GiGPO & \textbf{97.1} & 60.0 & \textbf{92.3} & \textbf{86.7} & 95.5 & \textbf{71.4} & \textbf{87.5} & 0.497B \\
% GSPO & 93.3 & \textbf{90.0} & 70.4 & 57.1 & 61.5 & 57.9 & 71.7 & 0.529B \\
% Tree-GRPO & 90.0 & 53.3 & 77.8 & 46.2 & 54.2 & 36.8 & 64.1 & 0.700B \\
% TriGRPO (ours) & 80.6 & 87.5 & 89.2 & 66.7 & \textbf{96.3} & 69.2 & 84.4 & \textbf{0.367B} \\
\bottomrule
\end{tabular}
\caption{Per-type and overall (\textbf{All}) peak validation SR (\%) on the ALFWorld \texttt{test} split, together with cumulative token consumption up to the peak step (\textbf{Tokens}). \textbf{Bold} marks the best value in each column (highest SR, lowest token cost). \textit{Base} is the untrained instruct model. }
\label{tab:alfworld}
\end{table*}

\subsection{ALFWorld: Long-Horizon Embodied Tasks}
\label{sec:alfworld_results}

Table~\ref{tab:alfworld} reports peak validation SR per task type and cumulative token consumption up to the peak step. TriGRPO attains the highest overall SR across all Qwen2.5 scales (88.3\%, 93.8\%, and 93.8\%), outperforming the strongest baseline at each scale by 1.6--6.2 points and root-only GRPO by 15.5--20.9 points. 
Notably, the margin over Tree-GRPO which shares prefix-reusing tree rollouts and differs in \emph{whether} branching is hindsight-guided by successful states, widens from 1.6 points at 1.5B to 16.5 points at 3B and 7B. These gains concentrate predominantly in multi-subgoal tasks requiring complex action composition: on \texttt{pick2} TriGRPO outperforms Tree-GRPO by 30--38 points at every scale, and on \texttt{heat} by over 40 points at 3B and 7B, while remaining competitive on shorter tasks. This directly reflects our mechanism: within-tree hindsight branching only triggers when an unguided rollout encounters a state matched in the success bank with a divergent action, providing targeted corrections precisely where failing trajectories diverge and where blind random expansion fails.

Beyond task performance, TriGRPO delivers substantial rollout efficiency. It is the most token-efficient method at 1.5B and 3B, requiring 32.5--39.4\% fewer tokens than Tree-GRPO and 14.5--16.1\% fewer than GRPO. At 7B, it remains second only to GiGPO (trailing by 0.054B tokens while outperforming it by 3.0 SR points). Complete cost comparisons, per-type localization, and baseline ordering across scales are detailed in Appendix~\ref{app:alfworld_details}.

\begin{table*}[ht]
\centering
\small
\begin{tabular}{l ccccc cccc}
\toprule
 & \multicolumn{5}{c}{\textbf{Multi-hop}} & \multicolumn{4}{c}{\textbf{Single-hop}} \\
\cmidrule(lr){2-6} \cmidrule(lr){7-10}
\textbf{Method} & HotpotQA & 2Wiki & MuSiQue & Bamboogle & \textbf{Avg.} & NQ & TriviaQA & PopQA & \textbf{Avg.} \\
\midrule
\multicolumn{10}{l}{\textit{Qwen2.5-1.5B-Instruct}} \\
\midrule
Base (no RL) & 5.9 & 4.3 & 2.6 & 8.0 & 5.2 & 7.1 & 22.4 & 9.9 & 13.1 \\
GRPO & 14.6 & 24.4 & 2.2 & 4.0 & 11.3 & 39.4 & 51.0 & 39.7 & 43.3 \\
GiGPO & 16.7 & 25.6 & 3.4 & \textbf{54.8} & 25.1 & 16.4 & 34.8 & 14.9 & 22.0 \\
GSPO & 15.8 & 23.7 & 2.5 & 4.8 & 11.7 & 36.8 & 48.9 & 37.3 & 41.0 \\
Tree-GRPO & 29.5 & 26.8 & 6.6 & 13.6 & 19.1 & 43.6 & 57.3 & 41.6 & 47.5 \\
TriGRPO (ours) & \textbf{37.9} & \textbf{32.7} & \textbf{10.8} & 24.8 & \textbf{26.6} & \textbf{45.1} & \textbf{60.4} & \textbf{43.3} & \textbf{49.6} \\
\midrule
\multicolumn{10}{l}{\textit{Qwen2.5-3B-Instruct}} \\
\midrule
Base (no RL) & 14.9 & 24.4 & 2.0 & 2.4 & 10.9 & 10.6 & 28.8 & 10.8 & 16.7 \\
GRPO & 39.0 & 36.3 & 15.2 & 36.8 & 31.8 & 44.4 & 58.0 & 42.5 & 48.3 \\
GiGPO & 36.9 & 37.0 & 12.6 & \textbf{64.1} & 37.6 & 42.0 & 59.5 & 42.4 & 48.0 \\
GSPO & 40.2 & 39.8 & 17.0 & 36.8 & 33.5 & 43.0 & 58.8 & 42.5 & 48.1 \\
Tree-GRPO & 42.4 & \textbf{43.7} & 17.8 & 43.3 & 36.8 & \textbf{46.8} & 59.7 & 43.6 & 50.0 \\
TriGRPO (ours) & \textbf{51.8} & 41.7 & \textbf{21.2} & 38.5 & \textbf{38.3} & 46.6 & \textbf{63.0} & \textbf{46.6} & \textbf{52.1} \\
% \midrule
% \multicolumn{10}{l}{\textit{Qwen2.5-7B-Instruct}} \\
% \midrule
% Base (no RL) & 18.3 & 25.0 & 3.1 & 12.0 & 14.6 & 13.4 & 40.8 & 14.0 & 22.7 \\
% GRPO & 42.5 & 40.7 & 19.1 & 43.2 & 36.4 & 45.8 & 61.5 & 44.3 & 50.5 \\
% GiGPO & 41.6 & \textbf{43.6} & 18.9 & \textbf{68.9} & \textbf{43.3} & 46.4 & \textbf{64.7} & \textbf{46.1} & 52.4 \\
% GSPO & 40.0 & 38.2 & 19.2 & 44.0 & 35.4 & 47.0 & 64.5 & \textbf{46.1} & \textbf{52.5} \\
% Tree-GRPO & \textbf{44.6} & 42.3 & \textbf{20.2} & 44.0 & 37.8 & \textbf{48.1} & 63.3 & 45.2 & 52.2 \\
% TriGRPO (ours) & -- & -- & -- & -- & -- & -- & -- & -- & -- \\
\bottomrule
\end{tabular}
\caption{SearchQA accuracy (\%) on four multi-hop and three single-hop benchmarks, with \textbf{Avg.} the unweighted mean over each group. \textbf{Bold} marks the best value per column within a model block; \textit{Base} is the untrained instruct model. }
\label{tab:qa}
\end{table*}

\begin{table*}[t]
\centering
\small
\begin{tabular}{lcccccccc}
\toprule
\textbf{Method} & pick & look & clean & heat & cool & pick2 & \textbf{All} & \textbf{Tokens} \\
\midrule
TriGRPO (root+cross+node) & 96.6 & 86.7 & 95.8 & 100.0 & 91.3 & 88.2 & 93.8 & 0.338B \\
root+cross (w/o node) & 98.9 & 73.2 & 91.4 & 96.7 & 78.7 & 91.0 & 89.8 & 0.294B \\
root+node (w/o cross) & 93.1 & 70.8 & 90.0 & 69.9 & 71.4 & 80.2 & 81.9 & 0.249B \\
GiGPO+cross & 95.0 & 74.6 & 94.0 & 89.8 & 83.1 & 87.8 & 89.5 & 0.459B \\
GRPO (root only) & 91.4 & 63.6 & 70.6 & 71.4 & 64.0 & 76.2 & 72.9 & 0.403B \\
\bottomrule
\end{tabular}
\caption{Component ablation on ALFWorld (Qwen2.5-3B-Instruct, 150 training steps). Per-type and overall (\textbf{All}) peak validation SR (\%) are measured on the validation split ($n{=}128$, $t{=}0.4$); \textbf{Tokens} is the cumulative token consumption up to the peak step. The GiGPO+cross row replaces TriGRPO's hindsight-guided node grouping with GiGPO state-prefix grouping while retaining $A_{\text{cross}}$.}
\label{tab:ablation}
\end{table*}

\subsection{SearchQA: Retrieval-Augmented Multi-Hop QA}
\label{sec:qa_results}

In SearchQA the agent answers open-domain questions via iterative query-and-read steps, emitting a single scored answer and a sparse terminal reward per episode.
Following the prevailing mixed training recipe \citep{jin2025search}, we evaluate on seven benchmarks---four multi-hop (HotpotQA \citep{yang2018hotpotqa}, 2WikiMultihopQA \citep{ho2020constructing}, MuSiQue \citep{trivedi2022musique}, Bamboogle \citep{press2022measuring}) and three single-hop (Natural Questions \citep{kwiatkowski2019natural}, TriviaQA \citep{joshi2017triviaqa}, PopQA \citep{mallen2023not}); Table~\ref{tab:qa} summarizes the results.

TriGRPO achieves the top multi-hop averages at both scales (26.6\% at 1.5B, 38.3\% at 3B), outperforming GRPO by 15.3 and 6.5 points and the strongest baseline (GiGPO) by 1.5 and 0.7 points.
On such sparse-reward prompts every rollout can fail and the group-relative advantage vanishes; both forms of hindsight credit act: the across-tree signal $A_{\text{cross}}$ restores a nonzero, difficulty-normalized signal from the historical baseline of Eq.~\eqref{eq:ema-baseline}, while within-tree hindsight branching imports a query formulation that succeeded elsewhere in the batch into a failing trajectory at a matched decision state, supervised by $A_{\text{node}}$ rather than an uninformative group mean. Per-benchmark margins are in Appendix~\ref{app:qa_details}.

Single-hop benchmarks need only one retrieved evidence span rather than multi-round reasoning; here TriGRPO again leads with 49.6\% and 52.1\% average success over the strongest baseline (Tree-GRPO, 47.5\% and 50.0\%), topping TriviaQA and PopQA at both scales and NQ at 1.5B. Its margins over GRPO are about half the multi-hop ones ($+6.3$ and $+3.8$ vs.\ $+15.3$ and $+6.5$ points), consistent with the mechanism: with a single decisive span, unguided rollouts already solve enough prompts that within-group variance rarely collapses and within-tree hindsight branching fires less often, leaving few failing trajectories at matched states to receive an imported query formulation. Per-benchmark columns, including the small-set Bamboogle column, are in Appendix~\ref{app:qa_details}.

\subsection{Ablation Study}
\label{sec:ablation}

We conduct three complementary ablations on ALFWorld with Qwen2.5-3B-Instruct: a \emph{component ablation} that removes or replaces individual terms of the three-level advantage estimator, a \emph{prefix-credit ablation} that varies whether subtree credit is written into the loss of shared prefix tokens, and a \emph{cross-advantage ablation} that dissects the historical baseline of $A_{\text{cross}}$. Detailed protocols and the full results of the latter two suites are provided in Appendix~\ref{app:ablation_details}; here we present the component ablation and the key findings.

\paragraph{Component ablation.}
The full model combines root advantage $A_{\text{root}}$ (within-prompt group-relative), cross-sample advantage $A_{\text{cross}}$ (across-tree hindsight baseline), and node advantage $A_{\text{node}}$ (within-tree hindsight branch credit). We selectively remove or replace individual components while keeping all other configurations identical: \textbf{root+cross} (w/o node) sets $w_{\text{node}}{=}0$; \textbf{root+node} (w/o cross) sets $\lambda_{\text{cross}}{=}0$; and \textbf{GiGPO+cross} replaces within-tree hindsight node grouping with GiGPO's state-prefix grouping while retaining $A_{\text{cross}}$.

\paragraph{Results.}
Table~\ref{tab:ablation} summarizes the component ablation. The full TriGRPO achieves 93.8\% peak SR. Removing $A_{\text{node}}$ (root+cross) drops performance to 89.8\% ($-4.0$ points); removing $A_{\text{cross}}$ (root+node) drops it to 81.9\% ($-11.9$ points); and replacing hindsight branching with GiGPO state-prefix grouping (GiGPO+cross) yields 89.5\% ($-4.3$ points). All tree-structured configurations nevertheless outperform root-only GRPO (72.9\%) by 9.0--20.9 points, confirming that the root-level tree structure itself contributes substantially over flat GRPO sampling.

\paragraph{Analysis.}
Two findings emerge from the component ablation. \emph{(1)~Across-tree hindsight credit is the dominant contributor:} the 11.9-point drop when $\lambda_{\text{cross}}{=}0$ shows that $A_{\text{cross}}$ provides critical learning signal beyond within-prompt comparisons, sustaining informative gradients even when all rollouts for a prompt yield identical outcomes---a frequent failure mode in sparse-reward environments where GRPO's group-relative signal vanishes. \emph{(2)~Within-tree hindsight branch credit is not interchangeable with generic step-level grouping:} disabling $A_{\text{node}}$ costs 4.0 points (89.8\% vs.\ 93.8\%) and substituting GiGPO's state-prefix grouping under the same cross advantage costs 4.3 points (89.5\%), indicating that the gain comes from hindsight-guided branching itself---counterfactual branch points at matched states---rather than from the mere presence of any node-level grouping. The prefix-credit and cross-advantage ablations (findings~3--4) and the token-efficiency analysis are reported in Appendix~\ref{app:ablation_details}, specifically Appendix~\ref{app:prefix_credit}, \ref{app:cross_ablation}, and~\ref{app:token_efficiency}.

\subsection{Performance across Difficulty Levels}
\label{sec:hard_examples}

\begin{figure}[ht]
    \centering
    \includegraphics[width=\columnwidth]{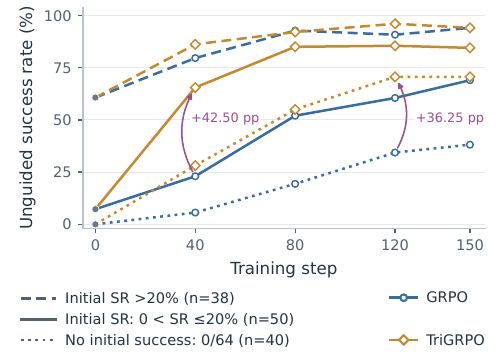}
    \par\vspace{2pt}
    {\footnotesize (a) Learning by initial difficulty\par}
    \vspace{8pt}
    \includegraphics[width=\columnwidth]{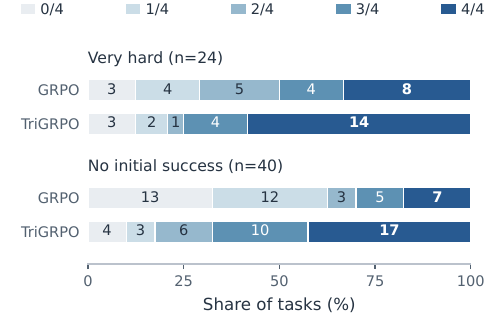}
    \par\vspace{2pt}
    {\footnotesize (b) Repeated unguided success\par}
    \caption{Autonomous performance by initial task difficulty. (a) Mean unguided success rates throughout training; annotations show TriGRPO--GRPO differences in percentage points. Initial trials define both the groups and the shared starting points. (b) Per-task success counts across four trials at step 150, for very hard tasks ($0<\hat p_0\leq0.05$) and tasks with no initial success. Segment widths show task proportions; numbers give task counts.}
    \label{fig:hard_examples}
\end{figure}

We examine whether TriGRPO improves autonomous task solving when initial successes are scarce. Starting from Qwen2.5-3B-Instruct, we train TriGRPO and root-only GRPO for 150 steps and evaluate on 128 held-out ALFWorld \texttt{valid\_seen} tasks. We estimate each task's initial success rate $\hat p_0$ from 64 unguided trials and freeze its difficulty label before training. For visualization, we group tasks into $\hat p_0>0.2$ (38 tasks), $0<\hat p_0\leq0.2$ (50 tasks), and no observed initial success (40 tasks). At steps 40, 80, 120, and 150, each task receives four unguided evaluation trials.

\paragraph{Learning dynamics.}
Figure~\ref{fig:hard_examples}(a) shows larger gains on initially difficult tasks. In the low but nonzero initial-success group, TriGRPO reaches 65.5\% success at step 40, compared with 23.0\% for GRPO, a gain of 42.5 percentage points. In the group with no initial success, the gap reaches 36.25 points at step 120 (70.6\% versus 34.4\%). The advantage persists at step 150: TriGRPO achieves 84.5\% versus 69.0\% in the former group and 70.6\% versus 38.1\% in the latter. Both methods reach 94.1\% in the highest-initial-success group, where less room for improvement remains. These trajectories indicate that TriGRPO acquires autonomous solving behavior earlier on tasks that are initially difficult for the shared starting policy.

\paragraph{Repeated success.}
Figure~\ref{fig:hard_examples}(b) distinguishes solving a task at least once from succeeding across repeated attempts. Among the 24 very hard tasks, both methods solve 21 at least once, but the number solved in all four trials rises from 8 with GRPO to 14 with TriGRPO. Among the 40 tasks with no initial success, the number solved at least once increases from 27 to 36, while the number solved in all four trials increases from 7 to 17. TriGRPO therefore improves both observed coverage and repeated success on the hardest group. Because all evaluations are unguided, these gains concern the learned policy's autonomous behavior and support the motivation of consolidating scarce successes into stronger task-solving capability.

\section{Conclusion}
% We present TriGRPO, a tree-based reinforcement learning framework that jointly addresses inefficient exploration and coarse credit assignment in long-horizon agent tasks. By coupling success-guided hindsight branching with hierarchical advantage estimation, TriGRPO transforms sparse successful experiences into targeted exploration opportunities and informative learning signals. Experiments demonstrate consistent performance gains and improved rollout efficiency across model scales. These findings highlight the value of integrating exploration and credit assignment, enabling agents to systematically leverage their own successes for more effective long-horizon learning.
We propose TriGRPO, a tree-based policy optimization method that performs hindsight credit assignment over trees via a triple-group structure. 
Empirical results across diverse tasks and model scales demonstrate superior performance and token efficiency, validating the effectiveness of online hindsight credit assignment.

\section*{Limitations}
% Our approach has three main limitations. First, hindsight branching relies on exact state hashing and intermediate environment restoration via snapshots or replay, which we do not evaluate in continuous, multimodal, or non-resettable environments. Second, branching requires at least one successful rollout for the same prompt to seed the bank, leaving unsolved prompts dependent solely on the cross-sample baseline. Finally, the historical baseline relies on rule-parsed task identifiers rather than learned representations, and we leave combinations with offline hindsight relabeling to future work.
%Our method has a primary limitation. Hindsight branching relies on exact state description matching and intermediate environment resettability, which may not readily extend to continuous or non-reversible domains. 
A key limitation of our approach lies in hindsight branching, which hinges on exact state matching and intermediate environment resettability. Consequently, extending this mechanism to continuous or irreversible domains remains challenging.

% Bibliography entries for the entire Anthology, followed by custom entries
%\bibliography{custom,anthology-overleaf-1,anthology-overleaf-2}

% Custom bibliography entries only
% Include only works cited in the manuscript; retain unused entries in custom.bib.
\bibliography{custom}

\appendix

\section{Detailed Experimental Settings}

\subsection{Environments and Tools}
\label{app:tools}
All models are trained with vLLM-accelerated tree rollouts. On ALFWorld we set the vLLM GPU memory fraction to 0.55 with tensor parallelism 1; on SearchQA the GPU memory fraction is 0.5 with tensor parallelism 1.

All experiments adopt the Qwen2.5-Instruct base model. Specifically, we evaluate on:
\begin{itemize}[leftmargin=*]
    \item \textbf{ALFWorld}: Evaluated across 1.5B, 3B, and 7B scales. It consists of six household interactive task types (\emph{pick, look, clean, heat, cool, pick-two}) evaluated on the standard $n=128$ test games.
    \item \textbf{SearchQA}: Evaluated on 1.5B and 3B scales. It comprises 51{,}713 multi-hop reasoning questions aggregated across seven standard open-domain QA test sets, requiring iterative web search and multi-turn reasoning.
\end{itemize}

\subsection{Hyperparameter Settings}
\label{app:hyperparams}

\paragraph{ALFWorld.}
All ALFWorld runs use the following hyperparameters unless otherwise noted. The tree rollout employs $R{=}8$ root trajectories per prompt with up to $L{=}12$ maximum leaves. We set the learning rate to $1\times10^{-6}$ with a constant schedule, KL penalty coefficient $\beta{=}0.01$ (low-variance estimator), and advantage clipping at 10.0. The training batch size is 16 prompts per step; the validation batch is 128 tasks. Maximum prompt and response lengths are 2048 and 512 tokens, respectively. For the across-tree hindsight advantage, we use scaling weight $\lambda_{\text{cross}}{=}1.2$, EMA smoothing $\alpha{=}0.3$, and shrinkage $\kappa{=}8.0$. The node-level advantage weight is $w_{\text{node}}{=}1.0$ with branch credit minimum gap 1.0. Within-tree hindsight branching uses an injection budget of 96, saturation threshold 0.5, divergence requirement enabled, with at most 2 branches per path and minimum branch interval 5. Behavior cloning applies weight 0.15 with a hold schedule (steps 30--100, floor 0.002). Training runs for 150 steps with evaluation every 5 steps, yielding 31 validation points.

\paragraph{SearchQA.}
On SearchQA the tree budget is reduced to $R{=}3$ root trajectories with up to $L{=}5$ leaves per prompt to accommodate longer retrieval-augmented contexts, and the training batch is 256 prompts per step, giving a rollout group of $n{=}5$ trajectories per prompt at temperature $t{=}1.0$. The environment horizon is 10 interaction steps with matching history length; maximum prompt and response lengths are 9216 and 512 tokens, respectively, and each retrieved observation is capped at 768 tokens. The learning rate ($1\times10^{-6}$, constant), advantage clipping (10.0), the across-tree hindsight weights ($\lambda_{\text{cross}}{=}1.2$, $\alpha{=}0.3$, $\kappa{=}8.0$), and the node-level weight ($w_{\text{node}}{=}1.0$) are as on ALFWorld, whereas the KL coefficient is $\beta{=}0.001$ under the same low-variance estimator and the branch credit minimum gap is lowered to 0.1. Within-tree hindsight branching keeps the injection budget of 96 and at most 2 branches per path, with the saturation threshold at 1.0, the divergence requirement disabled, and the minimum branch interval reduced to 1; the behavior cloning schedule (weight 0.15, steps 30--100, floor 0.002) is unchanged. Training follows a one-epoch budget of 663 update steps, with checkpoints selected by periodic validation every 10 steps on a fixed set of 400 multi-hop questions (100 per multi-hop benchmark, seed 42) and training stopped at the validation peak.

\subsection{Evaluation Protocols}
\label{app:eval_protocol}

\paragraph{Multi-Seed Validation.}
To ensure statistical robustness, all training runs are repeated across three independent random seeds. We record the peak validation success rate (SR) for each run and report the mean.

\paragraph{Sampling Efficiency Metric.}
We track cumulative token consumption up to the training step where the policy attains its peak validation SR. This reflects the token and compute budget required to reach optimal performance.

\paragraph{ALFWorld.}
We report peak validation success rate (SR) over the 31 evaluation checkpoints, measured on the ALFWorld \texttt{test} split ($n{=}128$) with temperature $t{=}0.4$ and sampling enabled. Per-type SR is reported for descriptive analysis; due to small per-type sample sizes (5--29), we do not perform statistical inference on individual task categories. We additionally report cumulative token consumption (\texttt{total\_num\_tokens}) up to the peak step as a measure of sampling efficiency.

\paragraph{SearchQA.}
Periodic validation is used only for checkpoint selection. Every reported SearchQA number comes from a separate full evaluation of the selected checkpoint: greedy decoding with an environment horizon of 10 steps on the complete test split of each benchmark (51{,}713 questions in total across the seven benchmarks).

\subsection{Baseline Reproduction}
\label{app:baselines}
All baselines are trained from the same Qwen2.5-Instruct checkpoints under the identical configuration and evaluation protocol described above (same rollout budget, decoding settings, and step count), so observed differences reflect the optimization objective rather than baseline-specific tuning.

\section{Experimental Details and Supplementary Results}

\subsection{ALFWorld Per-Type and Efficiency Details}
\label{app:alfworld_details}

\paragraph{Per-type profile.}
On the four shorter types (\texttt{pick}, \texttt{look}, \texttt{clean}, \texttt{cool}), TriGRPO is the best or tied-best method on five of the twelve scale-by-type cells---\texttt{clean} at 1.5B (100.0\%), 3B (95.8\%), and 7B (100.0\%, tied with GSPO), \texttt{look} at 3B (86.7\%), and \texttt{pick} at 7B (100.0\%, tied with GSPO)---and lies at or within three points of the best method on six of the twelve in total. It falls behind on the other six: by more than five points on \texttt{cool} at 1.5B (76.5 vs. Tree-GRPO's 95.2), \texttt{cool} at 7B (76.5 vs. GSPO's 96.8), \texttt{look} at 1.5B (66.7 vs. Tree-GRPO's 73.3), and \texttt{look} at 7B (94.4 vs. GSPO's 100.0), and by 3.9 and 4.8 points on \texttt{cool} at 3B (91.3 vs. GiGPO's 95.2) and \texttt{pick} at 1.5B (92.3 vs. Tree-GRPO's 97.1).

\paragraph{Why the margin widens with scale.}
The widening is not a scale effect in TriGRPO's favor alone: Tree-GRPO falls from 86.7\% overall SR at 1.5B to 77.3\% at 3B and 7B while consuming the largest token budget of any method (0.491--0.583B): its lead over flat GRPO shrinks to 4.4 points at 3B (77.3\% vs. 72.9\%) and reverses at 7B (77.3\% vs. 77.6\%). Undirected branching thus turns harmful as the policy strengthens---expanding random non-leaf nodes increasingly duplicates continuations the policy would already produce---whereas gating branches on a banked verified-successful action (Eq.~\eqref{eq:disagreement_gate}) keeps the same tree machinery improving with scale. Among the baselines the leader changes with scale---GiGPO and Tree-GRPO tie at 1.5B (86.7\%), GiGPO leads at 3B (87.6\%), and GSPO at 7B (91.0\%)---so no baseline is a consistent reference point, while TriGRPO ranks first at all three.

\paragraph{Sampling cost comparison.}
Prefix sharing alone does not account for this: Tree-GRPO shares prefixes yet costs up to 1.6$\times$ more. Within-tree hindsight branching converts banked successful actions into high-value continuations that are likely to complete the task, instead of another exploratory dead end whose tokens are spent without producing usable relative signal.

\paragraph{Cross-family run.}
%To test generality beyond the Qwen2.5 family, we train Llama-3.2-3B-Instruct under an identical configuration. 
TriGRPO obtains 84.4\% overall against 87.5\% for GiGPO, the strongest baseline at this scale, while using 26.2\% fewer tokens (0.367B vs.\ 0.497B) and 47.6\% fewer than Tree-GRPO (0.700B). Its per-type profile is complementary rather than uniformly weaker: it attains the best \texttt{cool} SR (96.3\%) and the second-best \texttt{look} (87.5\%) and \texttt{clean} (89.2\%), and the 3.1-point overall deficit is driven almost entirely by \texttt{pick}, on which all other RL baselines score 83.3--97.1\%. With a single evaluation seed over 128 tasks, 3.1 points correspond to four tasks; we therefore read this run as parity in accuracy at substantially lower sampling cost rather than as a counterexample, and rest the main claim on the three Qwen2.5 scales, which agree in direction and magnitude.

\subsection{QA Per-Benchmark Details}
\label{app:qa_details}

\paragraph{In-domain and unseen benchmarks.}
Two of them (NQ and HotpotQA) also appear in the training mixture and are therefore in-domain for all trained methods; the remaining five are unseen.

\paragraph{Per-benchmark margins.}
It is also the only method placed in the top two on all seven benchmarks at 1.5B. The advantage is largest exactly where the reward is sparsest: on MuSiQue, whose questions require three to four retrieval hops, the untrained models score 2.0--2.6\%, GRPO fails to improve over the base at 1.5B (2.2\% vs.\ 2.6\%), and Tree-GRPO reaches 6.6\% and 17.8\%, whereas TriGRPO reaches 10.8\% and 21.2\%.

\paragraph{Tree-baseline ranking across environments.}
The two environments also rank the tree baseline differently: Tree-GRPO, which is the most expensive method on ALFWorld and falls below flat GRPO there at 7B (77.3\% vs.\ 77.6\%), is the strongest baseline on both single-hop QA averages. Undirected branching retains value when an action is an arbitrary query string rather than one of a handful of environment commands, so random expansion still yields diverse informative continuations. Even there it trails TriGRPO---by 2.1 points on the single-hop average at both scales and by 7.5 and 1.5 points on the multi-hop average---which separates the contribution of the branching criterion from that of the tree structure itself.

\paragraph{Small-set columns.}
Bamboogle contains only 125 test items, so a single question moves its score by 0.8 points, and it is the only benchmark on which TriGRPO falls outside the top two at either scale (24.8\% and 38.5\%, against GiGPO's 54.8\% and 64.1\%). GiGPO's Bamboogle scores coexist with single-hop accuracy below GRPO's at both scales (1.5B: NQ 16.4\% vs.\ 39.4\%, PopQA 14.9\% vs.\ 39.7\%), a profile more consistent with high-variance optimization on a small set than with uniform multi-hop skill. Apart from Bamboogle, TriGRPO's only non-leading columns at 3B are 2Wiki (41.7 vs.\ 43.7) and NQ (46.6 vs.\ 46.8), gaps of 2.0 and 0.2 points against Tree-GRPO. We therefore treat the group averages, where the ordering is stable across scales, as the primary comparison and read individual small-set columns as descriptive, in line with the protocol of Section~\ref{sec:experiment_setup}. 
% At 7B, where our run is still in progress, GiGPO leads the multi-hop average (43.3\%) and GSPO the single-hop average (52.5\%).

\section{Ablation Details}
\label{app:ablation_details}

\subsection{Component Ablation Groups}
\label{app:component_ablation}
\begin{itemize}
    \item \textbf{root + cross} (w/o node): Sets $w_{\text{node}}{=}0$, disabling within-tree hindsight branch credit. Unguided trajectories still receive both root and cross advantages; hindsight branches contribute only through behavior cloning.
    \item \textbf{root + node} (w/o cross): Sets $\lambda_{\text{cross}}{=}0$, removing the across-tree hindsight baseline. Trajectories receive root advantage plus within-tree hindsight branch credit, but without difficulty-normalized correction from past performance on related tasks.
    \item \textbf{GiGPO + cross}: Replaces TriGRPO's within-tree hindsight node grouping with GiGPO's state-prefix grouping, while retaining the same across-tree hindsight advantage ($\lambda_{\text{cross}}{=}1.2$). This tests whether the node-level signal requires hindsight branching or can be achieved through alternative step-level grouping mechanisms.
\end{itemize}

\subsection{Prefix-Credit Ablation}
\label{app:prefix_credit}
By default, TriGRPO enforces a credit-isolation invariant: only node-level credit enters the PPO/GRPO policy loss, and it multiplies exclusively the tokens (actions) added \emph{after} the branch point of the corresponding child; shared prefix tokens preceding the branch point neither gain nor lose gradient from the quality of the child branches, and subtree returns are never back-propagated or aggregated into ancestor or root advantages. We ablate this design with three variants that additionally write subtree credit into the loss of shared prefix tokens: (i)~\textbf{mean aggregation}, which injects the average performance of on-policy continuations sampled from the prefix under the current policy; (ii)~\textbf{max aggregation}, which injects the best continuation observed from that prefix, asking whether an exceptionally good completion ever existed; and (iii)~\textbf{shuffle}, a control that permutes the real per-prefix credits already computed within the current batch without replacement, preserving their mean, variance, extremes, and total gradient scale while breaking the correspondence between each credit and its prefix.

\begin{table*}[t]
\centering
\small
\begin{tabular}{lcccccccc}
\toprule
\textbf{Method} & pick & look & clean & heat & cool & pick2 & \textbf{All} & \textbf{Tokens} \\
\midrule
TriGRPO (prefix-only loss) & 96.6 & 86.7 & 95.8 & 100.0 & 91.3 & 88.2 & 93.8 & 0.338B \\
+ branch mean aggregation & 89.7 & 87.5 & 96.2 & 90.0 & 68.2 & 95.7 & 88.3 & 0.292B \\
+ branch max aggregation & 100.0 & 62.5 & 100.0 & 88.2 & 90.0 & 71.4 & 88.3 & 0.389B \\
+ branch mean agg.\ w/ shuffle & 94.3 & 80.0 & 95.0 & 73.3 & 73.9 & 96.0 & 87.5 & 0.325B \\
\bottomrule
\end{tabular}
\caption{Prefix-credit ablation on ALFWorld (Qwen2.5-3B-Instruct); columns follow Table~\ref{tab:ablation}. The default TriGRPO row writes only each trajectory's own outcome credit into its tokens (branch credit applies solely to tokens after the branch point). The mean variant additionally writes the average on-policy continuation performance from each shared prefix into the prefix tokens' loss; the max variant instead writes the best continuation observed from each prefix into the prefix tokens' loss; the shuffle control permutes the real per-prefix credits within each batch without replacement, preserving their mean, variance, extremes, and total gradient scale while breaking the credit--prefix correspondence.}
\label{tab:prefix_credit}
\end{table*}

Table~\ref{tab:prefix_credit} reports the prefix-credit ablation: writing subtree credit into shared prefix tokens degrades the full model from 93.8\% to 88.3\% under mean aggregation, to 88.3\% under max aggregation, and to 87.5\% under the shuffle control. \emph{(3)~Injecting subtree credit into shared prefixes is harmful, and the harm is neither a scaling artifact nor a content artifact.} All three variants that violate the credit-isolation invariant degrade the full model substantially (mean aggregation 88.3\%, max aggregation 88.3\%, shuffle 87.5\%, vs.\ 93.8\%), and the three remain within 0.8 points of one another. The shuffle control is diagnostic: it preserves the mean, variance, extremes, and total gradient scale of the injected credits while destroying only their correspondence to the originating prefixes, yet it degrades performance just as much as the faithful mean variant---so the damage stems from the act of charging shared prefix tokens for the quality of downstream subtrees, not from a mismatched credit scale. The max variant is diagnostic in the orthogonal direction: it injects the strongest possible positive signal (the best continuation ever observed from each prefix, an upper-biased version of the mean credit), yet it fails to beat mean aggregation, so the harm is not attributable to the pessimistic content of the injected signal either. Notably, the existence of an exceptionally good continuation from a prefix---the question the max variant was designed to answer---carries no exploitable learning signal once written into the prefix tokens. This supports TriGRPO's default credit isolation: prefix tokens receive only the outcome credit of their own rows, while branch credit acts exclusively on tokens generated after the branch point.

\subsection{Across-Tree Hindsight (Cross) Advantage Ablation}
\label{app:cross_ablation}
The across-tree hindsight advantage corrects each trajectory's return by a historical baseline $\mu[t]$ maintained per task $t$. We track two exponential moving averages, a fine-grained per-task baseline $\mu_{\text{fine}}[t]$ and a coarse-grained per-type baseline $\mu_{\text{coarse}}[c]$, each updated with smoothing coefficient $\alpha$ toward the current-batch mean return, and combine them by shrinkage,
\begin{equation}
\mu[t] \;=\; \frac{n_t\,\mu_{\text{fine}}[t] + \kappa\,\mu_{\text{coarse}}[c(t)]}{n_t + \kappa},
\end{equation}
where $n_t$ is the accumulated sample count of task $t$ and $\kappa$ is a prior equivalent sample size; the advantage is then $A_{\text{cross}} = (R - \mu[t])/\text{scale}$ with $\text{scale}{=}10$ normalizing ALFWorld's $0/10$ success reward to roughly $[-1,1]$. The default configuration uses $\alpha{=}0.3$ and $\kappa{=}8$. We ablate three aspects: (i)~\textbf{$\kappa{=}0$} (no coarse prior), which drops the task-type shrinkage term so the baseline relies purely on fine per-task history, testing whether the coarse prior is useful; (ii)~\textbf{no fine EMA} (coarse-only), which removes the fine per-task baseline and keeps only the task-type EMA through a dedicated coarse-only cross-advantage function, testing whether fine-grained task identity matters; and (iii)~\textbf{$\alpha{=}1.0$}, which replaces the smoothed history with the raw current-batch mean, testing whether historical smoothing is necessary.

\begin{table*}[t]
\centering
\small
\begin{tabular}{lcccccccc}
\toprule
\textbf{Method} & pick & look & clean & heat & cool & pick2 & \textbf{All} & \textbf{Tokens} \\
\midrule
TriGRPO ($\alpha{=}0.3$, $\kappa{=}8$) & 96.6 & 86.7 & 95.8 & 100.0 & 91.3 & 88.2 & 93.8 & 0.338B \\
fine-only  ($\kappa{=}0$) & 100.0 & 77.8 & 93.3 & 88.2 & 94.4 & 78.3 & 90.6 & 0.383B \\
coarse-only  (fine EMA off) & 96.8 & 87.5 & 94.6 & 91.7 & 81.5 & 92.3 & 91.4 & 0.375B \\
no EMA smoothing ($\alpha{=}1.0$) & 96.4 & 84.6 & 85.7 & 84.6 & 69.6 & 78.3 & 83.6 & 0.417B \\
\bottomrule
\end{tabular}
\caption{Cross-advantage ablation on ALFWorld (Qwen2.5-3B-Instruct); columns follow Table~\ref{tab:ablation}. All variants keep the full three-level estimator and modify only the historical baseline $\mu[t]$ of $A_{\text{cross}}$. The default row uses EMA smoothing $\alpha{=}0.3$ with coarse-prior shrinkage $\kappa{=}8$; $\kappa{=}0$ removes the task-type prior, ``coarse-only'' removes the fine per-task EMA, and $\alpha{=}1.0$ replaces the smoothed history with the raw current-batch mean.}
\label{tab:cross_ablation}
\end{table*}

Table~\ref{tab:cross_ablation} dissects the cross-sample baseline: relative to the default two-component configuration ($\alpha{=}0.3,\kappa{=}8$; 93.8\%), fine-only (keeping only the fine per-task EMA) costs 3.2 points (90.6\%), coarse-only (keeping only the coarse prior) costs 2.4 points (91.4\%), and disabling EMA smoothing ($\alpha{=}1.0$) costs 10.2 points (83.6\%). \emph{(4)~The across-tree hindsight baseline needs both historical smoothing and its two-level fine/coarse structure.} Disabling EMA smoothing ($\alpha{=}1.0$) is by far the most damaging single change (83.6\%, $-10.2$ points): substituting the raw current-batch mean for the smoothed history makes the baseline high-variance and non-stationary, so the difficulty-normalized correction that sustains informative gradients on sparse-reward prompts collapses. The two-level structure is likewise justified---removing the coarse task-type prior ($\kappa{=}0$, $-3.2$ points) and removing the fine per-task EMA (coarse-only, $-2.4$ points) both degrade performance. The fine baseline captures task-specific difficulty while the coarse prior supplies a stable fallback when fine-grained samples are scarce, and their shrinkage combination is strictly better than either component alone.

\subsection{Token Efficiency}
\label{app:token_efficiency}
Tree-structured configurations consume 0.249--0.338B tokens (full TriGRPO 0.338B; root+node 0.249B), whereas the flat-sampling alternatives require 0.403B (GRPO) and 0.459B (GiGPO+cross)---a token reduction of roughly 16--38\% for tree rollouts. Within the prefix-credit ablation, the mean (0.292B) and shuffle (0.325B) variants are cheaper than the full model yet perform 5.5 and 6.3 points worse, while the max variant is the most expensive configuration of the entire study (0.389B, above even the full model's 0.338B) and still scores 5.5 points lower---showing that violating credit isolation buys neither accuracy nor sampling efficiency. The same holds for the cross-advantage baseline: the default $\alpha{=}0.3,\kappa{=}8$ configuration is also the most token-efficient (0.338B), whereas every ablated variant consumes more tokens (0.375--0.417B) yet scores lower, making the default cross configuration Pareto-optimal on both success rate and sampling cost.

\section{Prompts}\label{app:prompt}

\textbf{Prompt sources and parity.}
All compared methods (our \textbf{TriGRPO} and the baselines: \textbf{GRPO}, \textbf{GiGPO}, and \textbf{Tree-GRPO}) share exactly the same task prompt templates.
Only the advantage estimator and the rollout expansion policy differ, so any measured gap cannot be attributed to prompt engineering. The templates are defined in the environment package of our verl codebase, and follow the original schema of each task:

\begin{itemize}
  \item \textbf{ALFWorld (embodied agent):} follows the standard ALFWorld (ALFRED)
  text-environment ReAct schema. The agent reasons inside
  \texttt{<think>} and selects one admissible action inside \texttt{<action>}.
 
  \item \textbf{QA (search agent):} follows the search-augmented ReAct schema
  introduced by Search-R1. The agent reasons inside
  \texttt{<think>}, optionally issues a retrieval query inside \texttt{<search>},
  and finally emits the answer inside \texttt{<answer>}.
  
\end{itemize}

\textbf{Multi-turn variants.}
Both tasks are multi-turn. At the first turn there is no history yet, so a
no-history template is used (Figures~\ref{fig:alfworld_prompt_nohis}
and~\ref{fig:qa_prompt_nohis}); from the second turn onward the prompt is augmented
with the recent interaction history (Figures~\ref{fig:alfworld_prompt}
and~\ref{fig:qa_prompt}). The two variants differ only in the presence of the
history block; the task description, the available actions/tools, and the output
format instructions are identical. Both variants are shared verbatim by TriGRPO and
all baselines.

\textbf{Hindsight-branch prompt.}
The only TriGRPO-specific change to the model input occurs on hindsight branches, and only at their first turn.
To seed such a branch with a previously successful action, one successful action is injected into the generation prompt, as illustrated in Figure~\ref{fig:guided_forced_prompt}. 
The observation content itself (task, observation, admissible actions/queries) is left unchanged, and no answer, reward, or future observation is ever injected. 
The injected action span is subsequently relocated to the response side and supervised only by the behavior-cloning (BC) loss; the remaining turns of the branch are generated autonomously. Unguided rollouts, as well as all baseline roll-outs, never receive this prefix.

% ================= ALFWorld (embodied agent), NO-HIS =================
\begin{figure*}[htbp]
\begin{tcolorbox}[
    colback=gray!20,  % 20% gray,
    colframe=black,
    ] % title=
You are an expert agent operating in the ALFRED Embodied Environment. \\
Your current observation is: \{current\_observation\} \\
Your admissible actions of the current situation are: [\{admissible\_actions\}].

Now it's your turn to take an action. \\
You should first reason step-by-step about the current situation. This reasoning process MUST be enclosed within \texttt{<think>} \texttt{</think>} tags. \\
Once you've finished your reasoning, you should choose an admissible action for current step and present it within \texttt{<action>} \texttt{</action>} tags.
\end{tcolorbox}
\caption{First-turn (no-history) prompt template for the ALFWorld embodied agent.}
\label{fig:alfworld_prompt_nohis}
\end{figure*}

% ================= QA (search agent), NO-HIS =================
\begin{figure*}[htbp]
\begin{tcolorbox}[
    colback=gray!20,  % 20% gray,
    colframe=black,
    ] % title=
You are an expert agent tasked with answering the given question step-by-step. \\
Your question: \{task\_description\}

Now it's your turn to respond for the current step. \\
You should first conduct reasoning process. This process MUST be enclosed within \texttt{<think>} \texttt{</think>} tags. \\
After completing your reasoning, choose only one of the following actions (do not perform both): \\
(1) If you find you lack some knowledge, you can call a search engine to get more external information using format: \texttt{<search>} your query \texttt{</search>}. \\
(2) If you have enough knowledge to answer the question confidently, provide your final answer within \texttt{<answer>} \texttt{</answer>} tags, without detailed illustrations. For example, \texttt{<answer>Beijing</answer>}.
\end{tcolorbox}
\caption{First-turn (no-history) prompt template for the search QA agent.}
\label{fig:qa_prompt_nohis}
\end{figure*}

% ================= ALFWorld (embodied agent) =================
\begin{figure*}[htbp]
\begin{tcolorbox}[
    colback=gray!20,  % 20% gray,
    colframe=black,
    ] % title=
You are an expert agent operating in the ALFRED Embodied Environment. Your task is to: \{task\_description\} \\
Prior to this step, you have already taken \{step\_count\} step(s). Below are the most recent \{history\_length\} observations and the corresponding actions you took: \{action\_history\} \\
You are now at step \{current\_step\} and your current observation is: \{current\_observation\} \\
Your admissible actions of the current situation are: [\{admissible\_actions\}].

Now it's your turn to take an action. \\
You should first reason step-by-step about the current situation. This reasoning process MUST be enclosed within \texttt{<think>} \texttt{</think>} tags. \\
Once you've finished your reasoning, you should choose an admissible action for current step and present it within \texttt{<action>} \texttt{</action>} tags.
\end{tcolorbox}
\caption{Prompt template for the ALFWorld embodied multi-turn agent.}
\label{fig:alfworld_prompt}
\end{figure*}

% ================= QA (search agent) =================
\begin{figure*}[htbp]
\begin{tcolorbox}[
    colback=gray!20,  % 20% gray,
    colframe=black,
    ] % title=
You are an expert agent tasked with answering the given question step-by-step. \\
Your question: \{task\_description\}

Prior to this step, you have already taken \{step\_count\} step(s). Below is the interaction history where \texttt{<search>} \texttt{</search>} wrapped your past search queries and \texttt{<information>} \texttt{</information>} wrapped the corresponding search results returned by the external search engine. History: \\
\{memory\_context\}

Now it's your turn to respond for the current step. \\
You should first conduct reasoning process. This process MUST be enclosed within \texttt{<think>} \texttt{</think>} tags. \\
After completing your reasoning, choose only one of the following actions (do not perform both): \\
(1) If you find you lack some knowledge, you can call a search engine to get more external information using format: \texttt{<search>} your query \texttt{</search>}. \\
(2) If you have enough knowledge to answer the question confidently, provide your final answer within \texttt{<answer>} \texttt{</answer>} tags, without detailed illustrations. For example, \texttt{<answer>Beijing</answer>}.
\end{tcolorbox}
\caption{Prompt template for the search-augmented multi-turn QA agent.}
\label{fig:qa_prompt}
\end{figure*}

% ================= Guided branch: forced-action prefix =================
\begin{figure*}[htbp]
\begin{tcolorbox}[
    colback=gray!20,  % 20% gray,
    colframe=black,
    ] % title=
% --- ALFWorld (embodied action task) ---
[observation prompt as above for action task] \\
\texttt{<think></think><action>} \{forced\_action\} \texttt{</action>}

% --- QA (search task) ---
[observation prompt as above for QA task] \\
\texttt{<think></think><search>} \{forced\_query\} \texttt{</search>}
\end{tcolorbox}
\caption{Hindsight branch: at training time TriGRPO injects one successful action (empty \texttt{<think>} + a single action span) into the generation prompt of the first step of a hindsight branch. Observation content is unchanged; the injected span is relocated to the response side and receives only behavior-cloning (BC) loss.}
\label{fig:guided_forced_prompt}
\end{figure*}

\section{Code}\label{app:code}

TriGRPO is task-agnostic. 
The only task-dependent ingredients are (i) the \emph{decision state} that decides when two decisions occur at the same environment state, and (ii) the \emph{action surface} (a search query for QA, an admissible command for ALFWorld).
The two tasks expose different primitives, which we abstract behind this single descriptor: in ALFWorld the state is read directly from the simulator as the tuple ($gamefile$, $observation$, $sorted(admissible\_commands)$), whose hash uniquely identifies a room configuration together with the current inventory, so two decisions are at the same state iff their hashes match; in QA there is no simulator snapshot, and the state is the task together with the ordered \emph{chain of queries} already issued (the query prefix), so two decisions are at the same state iff they share the same question and query prefix.
Everything downstream is written against this single abstraction and is therefore shared verbatim across both tasks: the success bank $\mathcal{B}$ update and selection, the counterfactual branch created from a matched state, and the three-level advantage estimator.
Listing~\ref{list:state_bank} shows the decision-state definition and equality test, the success-bank update and selection, and the counterfactual branch; Listing~\ref{list:adv} shows the three-level advantage estimator with its cross-sample baseline.

\begin{figure*}
\begin{minipage}{0.95\textwidth}
\centering
\begin{lstlisting}[language=Python, caption={Decision-state definition and equality, the success bank $\mathcal{B}$, and the counterfactual branch. Only the decision state and the action surface differ between QA and ALFWorld; the bank policy is shared.}, label={list:state_bank}]
# ---- decision-state definition & equality (the ONLY task-dependent part) ----
# ALFWorld : state = (gamefile, observation, sorted(admissible_commands))
#            equal iff blake2b hash matches  -> same room/objects/inventory
# QA       : state = (question, ordered query prefix)
#            equal iff they share the same query prefix
def state_key(task, snap):
    if task == "alfworld":
        return blake2b(snap.gamefile, snap.observation,
                       sorted(snap.admissible_commands))
    return (task_fine_key(snap.source, snap.question),
            tuple(v.query for v in snap.visits))

# ---- per-prompt success bank $\mathcal{B}$ (task-agnostic update policy) ----
class SuccessBank:
    def register_success(self, traj_uid, visits, total_steps):
        # one candidate action per visit, ranked by (steps_to_success, uid, depth, action)
        for v in visits:
            if not v.valid:
                continue
            donor = DonorAction(v.action, total_steps - v.depth, traj_uid, v.depth)
            keep_best(self._bank, donor, key=donor_rank)   # dedup, keep best

    def select(self, seen, remaining_steps, accept):
        if self.saturated:                        # successes / roots >= ratio
            return None
        if remaining_steps < 2:
            return None
        for donor in sorted(self._bank.values(), key=donor_rank):
            if donor.action in seen:                    continue
            if donor.steps_to_success > remaining_steps: continue
            if not accept(donor.action):                continue
            return donor                               # best valid, unseen donor

# ---- reuse: on a matched state, create a counterfactual branch from the exact
#      pre-action snapshot and inject one successful donor action ----
donor = bank.select(state.seen_actions, remaining_steps, accept=action_fits)
if donor is not None:
    branch = parent.clone_at_matched_state(env.snapshot(state))  # same state
    branch.pending_inject_text = inject_target(donor.action)
    branch.pending_forced_action = True
\end{lstlisting}
\end{minipage}
\end{figure*}

\begin{figure*}
\begin{minipage}{0.95\textwidth}
\centering
\begin{lstlisting}[language=Python, caption={Three-level advantage estimator with the cross-sample baseline, shared by ALFWorld and QA. Each trajectory receives exactly one signal class: unguided trajectories get the root term plus the cross-sample baseline; guided branches get the node term.}, label={list:adv}]
# verl/trainer/ppo/core_algos.py -- identical structure for ALFWorld and QA

def ema_baseline(R, fine_key, coarse_key, state, cfg):
    # blend a per-task EMA (fine) with a per-task-type EMA (coarse) by shrinkage;
    # cold start falls back to whichever EMA exists, else 0
    mu_f = state.mu_fine.get(fine_key)          # None on first occurrence
    mu_c = state.mu_coarse.get(coarse_key)
    if mu_f is None and mu_c is None: return zeros_like(R)
    if mu_f is None:                  return mu_c
    if mu_c is None:                  return mu_f
    n = state.count_fine.get(fine_key, 0)       # #root trajectories seen (fine)
    return (n * mu_f + cfg.kappa * mu_c) / (n + cfg.kappa)   # kappa = 8

def update_ema(R, fine_key, coarse_key, state, alpha):      # alpha = 0.3
    # updated AFTER this step's advantages are computed (preserves causality)
    m = R.mean()
    state.mu_fine[fine_key]     = ema(state.mu_fine.get(fine_key),     m, alpha)
    state.mu_coarse[coarse_key] = ema(state.mu_coarse.get(coarse_key), m, alpha)
    state.count_fine[fine_key]  = state.count_fine.get(fine_key, 0) + 1
# ema(old, new, alpha) = alpha * new + (1 - alpha) * old   (first step: = new)

def compute_hindsight_tree_advantage(batch, cfg):
    R = batch.returns

    # (1) ROOT : GRPO over the unguided trajectories of the same prompt
    A_root = group_relative(R, group=batch.uid)
 
    # (2) NODE : outcome difference within each 2-member node group
    #            (the parent trajectory + the guided branch cloned from its state)
    A_node = batch.R_branch - batch.R_parent        # no variance normalization
 
    # (3) CROSS: deviation of the return from a historical baseline
    A_cross = (R - ema_baseline(R, batch.fine_key, batch.coarse_key,
                                batch.state, cfg)) / cfg.sigma_r
 
    return where(batch.tree_guided,
                 cfg.lambda_node * A_node,
                 A_root + cfg.lambda_cross * A_cross)
\end{lstlisting}
\end{minipage}
\end{figure*}

\end{document}